\documentclass[conference]{IEEEtran}
\IEEEoverridecommandlockouts
\usepackage{cite}
\usepackage{amsmath,amssymb,amsfonts}
\usepackage{algorithmic}
\usepackage{graphicx}
\usepackage{textcomp}

\usepackage{adjustbox} 
\usepackage{tabulary}
\usepackage{multirow}
\usepackage{ctable}
\usepackage{bbding} % Checkmark, XSolid 등 제공
\usepackage[hyphens]{url}  % DO NOT CHANGE THIS

\def\BibTeX{{\rm B\kern-.05em{\sc i\kern-.025em b}\kern-.08em
    T\kern-.1667em\lower.7ex\hbox{E}\kern-.125emX}}
\begin{document}

\title{LED: A Benchmark for Evaluating Layout Error Detection in Document Analysis

% *\\
% { \textsuperscript{*}Note: Sub-titles are not captured in Xplore and
% should not be used}
% \thanks{Identify applicable funding agency here. If none, delete this.}
}

\author{\IEEEauthorblockN{1\textsuperscript{st} Inbum Heo}
\IEEEauthorblockA{\textit{\textit{Department of}} \\
\textit{Computer Engineering}\\
\textit{Chungnam National Univ.}\\
Daejeon, South Korea \\
inbum1025@o.cnu.ac.kr}
\and
\IEEEauthorblockN{2\textsuperscript{nd} Taewook Hwang}
\IEEEauthorblockA{\textit{\textit{Department of}} \\
\textit{Computer Engineering}\\
\textit{Chungnam National Univ.}\\
Daejeon, South Korea \\
taewook5295@gmail.com}
\and
\and
\IEEEauthorblockN{3\textsuperscript{nd} Jeesu Jung}
\IEEEauthorblockA{\textit{\textit{Department of}} \\
\textit{Computer Engineering}\\
\textit{Chungnam National Univ.}\\
Daejeon, South Korea \\
jisu.jung5@gmail.com}
\and
\IEEEauthorblockN{4\textsuperscript{th} Sangkeun Jung\textsuperscript{\dag}}
\IEEEauthorblockA{\textit{\textit{Department of}} \\
\textit{Computer Engineering}\\
\textit{Chungnam National Univ.}\\
Daejeon, South Korea\\
hugmanskj@gmail.com
}
\thanks{\textsuperscript{\dag}Corresponding author.}
\thanks{\textsuperscript{*}This work was supported by Chungnam National University and by the National Research Foundation of Korea (NRF) grant funded by the Korea government (MSIT) (No. RS-2025-0055621731482092640101).}
}

\maketitle

% \begin{abstract}
% Recent advances in large language models (LLMs) and multimodal models have significantly improved the performance of Document Layout Analysis (DLA). However, structural errors such as region merging, splitting, and omission remain unresolved challenges. Conventional overlap-based metrics such as IoU and mAP have limitations in capturing these errors. Therefore, this study systematically defines eight standard error types that occur in document layout analysis and proposes a new benchmark, Layout Error Detection (LED), to evaluate them. LED consists of a synthetic dataset, LED-Dataset, constructed to reflect the real-world error distribution of models, and three evaluation tasks—document-level error detection, document-level error type classification, and element-level error type classification. Experiments with various large-scale multimodal models demonstrate that LED enables quantitative evaluation of a model’s structural understanding of documents.
% \end{abstract}

\begin{abstract}

Recent advances in Large Language Models (LLMs) and Large Multimodal Models (LMMs) have improved Document Layout Analysis (DLA), yet structural errors such as region merging, splitting, and omission remain persistent.
Conventional overlap-based metrics (e.g., IoU, mAP) fail to capture such logical inconsistencies.
To overcome this limitation, we propose \textit{Layout Error Detection (LED)}, a benchmark that evaluates structural reasoning in DLA predictions beyond surface-level accuracy.
LED defines eight standardized error types (Missing, Hallucination, Size Error, Split, Merge, Overlap, Duplicate, and Misclassification) and provides quantitative rules and injection algorithms for realistic error simulation.
Using these definitions, we construct LED-Dataset and design three evaluation tasks: document-level error detection, document-level error-type classification, and element-level error-type classification.
Experiments with state-of-the-art multimodal models show that LED enables fine-grained and interpretable assessment of structural understanding, revealing clear weaknesses across modalities and architectures.
Overall, LED establishes a unified and explainable benchmark for diagnosing the structural robustness and reasoning capability of document understanding models.

\end{abstract}


\section{Introduction}

Recent advancements in Large Language Models (LLMs) and Large Multimodal Models (LMMs) have greatly enhanced the capabilities of Document AI systems~\cite{b1,b2}.
As a result, document images are now extensively utilized in tasks such as information extraction and document-level question answering, which require an integrated understanding of both visual layout and logical structure.
To enable such comprehensive comprehension, a fundamental preprocessing stage—\textit{Document Layout Analysis (DLA)}—is indispensable~\cite{b11}.
DLA decomposes a document page into meaningful components such as text blocks, tables, and figures, thereby directly impacting the accuracy of downstream tasks including OCR, information extraction~\cite{b3,b4}, and question answering~\cite{b5,b6,b7}.

Despite rapid progress in object detection models~\cite{b8, b9, b10} and vision-language models (VLMs)~\cite{vlm}, DLA outputs still suffer from various types of errors. Beyond typical \textit{localization errors} (e.g., slight misalignments in bounding boxes), a more severe issue lies in \textit{structural errors}, where semantically distinct regions are incorrectly merged or split. These structural errors substantially hinder document understanding performance. However, conventional metrics such as IoU~\cite{iou} and mAP~\cite{map} are insufficient to detect or interpret such structural issues, as they mainly capture spatial overlaps without reflecting the logical consistency of layout predictions.

To address this gap, we introduce a new evaluation task that we define as \textbf{Layout Error Detection (LED)}, which systematically diagnoses structural errors in document layout predictions.
The LED benchmark consists of the following components:

\begin{enumerate}
    \item \textbf{Definition of DLA-specific errors:}  
    We define eight types of structural errors commonly observed in DLA outputs (e.g., \textit{Missing}, \textit{Merge}, \textit{Split}, \textit{Hallucination}, etc.). 
    Each error type is accompanied by quantitative thresholds and rule-based diagnostic criteria, establishing a unified framework for reproducible error analysis across DLA models. 
    This formulation generalizes prior object-detection error taxonomies to the document domain.

    \item \textbf{Synthetic dataset construction (LED-Dataset):}  
    We develop a synthetic benchmark dataset, \textit{LED-Dataset}, by injecting structural errors into DLA model predictions based on real-world error distributions. 
    Our rule-based error injection algorithm produces realistic and diverse erroneous layouts, reflecting the frequency and geometry of observed model failures. 
    This dataset serves as a controlled environment for reproducible, diagnostic evaluation of layout understanding.

    \item \textbf{LED benchmark:}  
    The LED framework includes three complementary tasks: (1) \textit{Document-level error detection}, which assesses whether a given layout contains any structural inconsistency; (2) \textit{Document-level error type classification}, which predicts the dominant error categories within a document; and (3) \textit{Element-level error classification}, which localizes and categorizes each erroneous region. 
    Together, these tasks provide a hierarchical perspective on how well a model can perceive, reason about, and differentiate structural irregularities within complex layouts.
\end{enumerate}

Using the LED benchmark, we evaluate a wide range of multimodal models to examine their structural understanding of documents.
LED goes beyond conventional accuracy-based evaluation by diagnosing where and why models fail, distinguishing those that merely detect layout errors from those that can reason about hierarchical and relational document structures.
By revealing systematic weaknesses across different modality configurations and architectures, LED establishes a new paradigm for explainable evaluation of model behavior, providing a foundation for developing more reliable and structurally aware document understanding systems.

\section{Related Work}

Existing studies in Document Layout Analysis (DLA) have aimed to segment and categorize document images into meaningful structural units.
Traditional approaches rely on detection-based frameworks (e.g., YOLO~\cite{bochkovskiy2020yolov4}, Deformable-DETR~\cite{zhu2021deformabledetr}) and large-scale datasets such as PubLayNet~\cite{zhong2019publaynet} and DocLayNet~\cite{pfitzmann2022doclaynet}, while recent vision-language models (e.g., LayoutLM~\cite{xu2020layoutlm} and its successors) extend DLA into the multimodal domain by jointly modeling textual, visual, and spatial features~\cite{xu2021layoutlmv2, huang2021docformer}.
Despite these advances, evaluation practices remain focused on detection accuracy rather than the structural understanding of document layouts.

In the broader field of Document Understanding, evaluation has typically focused on downstream tasks such as Optical Character Recognition (OCR), question answering~\cite{mathew2021docvqa} and information extraction.
These tasks are usually assessed with geometric metrics (IoU, mAP) that overlook logical or structural inconsistencies~\cite{tito2021visualmrc, li2022structurallm}.
Although post-processing methods (e.g., region merging or rule-based correction) refine predictions, they lack the capacity to diagnose or categorize errors affecting the coherence of document layouts~\cite{malerba2003layoutcorrection}.

In general object detection, diagnostic frameworks such as TIDE~\cite{bolya2020tide} and ObjectLab~\cite{popovic2022objectlab} decompose errors into interpretable types (e.g., classification or localization).
However, they are designed for natural images and fail to capture the hierarchical dependencies and semantic links unique to document layouts (e.g., captions, tables, and figures).
These gaps highlight the need for a diagnostic framework tailored to document-specific structure and reasoning.

Beyond traditional detection metrics, recent research has emphasized explainable and diagnostic evaluation in multimodal understanding.
Benchmarks such as VLMEvalKit~\cite{zhu2024vlmevalkit} and DocBench~\cite{fang2024docbench} assess reasoning and comprehension in vision-language models, but primarily address semantic or textual reasoning rather than structural fidelity.

We therefore propose the Layout Error Detection (LED) benchmark, which systematically defines document-specific structural error types and integrates them into the evaluation process.
LED quantifies detection failures and enables fine-grained analysis of structural inconsistencies across document domains.
Its companion dataset, LED-Dataset, captures realistic error patterns observed in model outputs, enabling controlled and reproducible evaluation of robustness and reasoning.

\section{Error in Document Layout Analysis}

In this paper, we define eight standardized error types to establish a unified foundation for evaluating structural robustness in Document Layout Analysis (DLA).
Each type represents a characteristic failure mode, defined by geometric and semantic criteria for consistent and reproducible comparison across models and domains.
The \textbf{LED-Dataset} visualizes realistic errors—such as Missing, Split, Merge, and Overlap—frequently observed in model outputs, enabling fine-grained, quantitative analysis and forming the basis for dataset generation and evaluation in the proposed \textbf{LED benchmark}.

\subsection{Document-Specific Error Definition and Injection}

Based on the actual outputs of DLA models, we define eight structural error types and their corresponding injection strategies.
Each error type is characterized by explicit structural patterns, discriminative rules, and injection conditions, enabling quantitative diagnosis and reproducible synthesis of realistic error scenarios.
To simulate these errors, we employ a rule-based injection algorithm that supports both document-level and element-level manipulation, as well as the composition of multiple error types within a single document.
In particular, Misclassification errors can co-occur with other structural errors, allowing diverse evaluation settings for robustness assessment.

Below, we describe each error type along with its corresponding injection procedure.

\begin{itemize}
  \item \textbf{Missing (False Negative):}  
A ground truth box $B_{\text{gt}}$ exists, but no predicted box $B_{\text{pred}}$ satisfies
  \[
    \text{IoU}(B_{\text{gt}}, B_{\text{pred}}) \geq 0.1
  \]
  \textit{Injection:} Approximately 10\% of ground truth annotations are randomly removed from the final annotation set, simulating false negatives where real objects are omitted.

  \item \textbf{Hallucination (False Positive):}  
A predicted box $B_{\text{pred}}$ exists, but no ground truth box $B_{\text{gt}}$ satisfies
  \[
    \text{IoU}(B_{\text{gt}}, B_{\text{pred}}) \geq 0.1
  \]
  \textit{Injection:} New boxes are inserted in regions without real objects, maintaining IoU $\leq 0.01$ with any existing ground truth boxes.

  \item \textbf{Size Error:}  
The predicted and ground truth boxes have similar centers, but their area ratio lies outside the acceptable range
  \[
    \frac{\text{area}(B_{\text{pred}})}{\text{area}(B_{\text{gt}})} \notin [0.6,\ 1.4]
  \]
  \textit{Injection:} Ground truth boxes are randomly scaled by 10–30\% around their centers while ensuring minimal overlap (IoU $\leq 0.01$) with neighboring boxes.

  \item \textbf{Split:}  
A single ground truth box $B_{\text{gt}}$ is fragmented into multiple predicted boxes $\{B_{\text{pred}}^{(i)}\}_{i=1}^{n}$ $(n \geq 2)$ such that
  {
  \[
    \forall i,\ \text{IoU}(B_{\text{gt}}, B_{\text{pred}}^{(i)}) < 0.5,\quad \sum_{i=1}^{n} \text{IoU}(B_{\text{gt}}, B_{\text{pred}}^{(i)}) \geq 0.5
  \]}
  \textit{Injection:} A single box is horizontally divided into $N$ narrow boxes with randomized width ratios, simulating over-segmentation errors.

  \item \textbf{Merge}  
\textit{Definition:} Two or more distinct ground truth boxes are merged into a single prediction $B_{\text{pred}}$, satisfying
{
\begin{align*}
  &\exists\, B_{\text{gt}}^{(i)} \neq B_{\text{gt}}^{(j)}\ \text{s.t.} \\
  &\text{IoU}(B_{\text{gt}}^{(i)}, B_{\text{pred}}) \geq 0.1 \\
  &\land\ \text{IoU}(B_{\text{gt}}^{(j)}, B_{\text{pred}}) \geq 0.1
\end{align*}
}

  \textit{Injection:} Spatially adjacent boxes (within 1.5× average width) belonging to the same class are merged into a single minimum enclosing rectangle.

  \item \textbf{Overlap:}  
Two predicted boxes overlap excessively
  \[
    \text{IoU}(B_i,\ B_j) \geq 0.1,\quad i \neq j
  \]
  \textit{Injection:} The center of an existing box is fixed, but its width and height are expanded to induce excessive overlap with nearby boxes.

  \item \textbf{Duplicate:}  
 Multiple predicted boxes correspond to the same ground truth object with high overlap
  \[
    \exists\, B_{\text{pred}}^{(1)}, B_{\text{pred}}^{(2)} : \text{IoU}(B_{\text{gt}}, B_{\text{pred}}^{(i)}) \geq 0.9
  \]
  \textit{Injection:} Several duplicate boxes are generated by perturbing the original box size by $\pm$10\% and shifting the center within a 10\% range.

  \item \textbf{Misclassification:}  
The predicted box correctly overlaps with the ground truth but has an incorrect label
  \[
    \text{IoU}(B_{\text{gt}}, B_{\text{pred}}) \geq 0.9,\quad \text{label}(B_{\text{gt}}) \neq \text{label}(B_{\text{pred}})
  \]
  \textit{Injection:} The predicted label is randomly replaced with another valid category within the dataset, reproducing semantic confusion between visually similar elements.
\end{itemize}

\begin{table}[!t]
\centering
\caption{Comparison between LED error types and existing error definition frameworks}
\begin{adjustbox}{max width=0.95\linewidth}
\begin{tabular}{cccccc}
\toprule 
\multirow{2}{*}{Error Type} & \multicolumn{5}{c}{Object Detection Error Type Definition Research}\tabularnewline
\cmidrule{2-6}
& LED (Ours) & TIDE & ObjectLab & Loss Inspection~\cite{schubert2024lossinspection} & DLER~\cite{vesalainen2024dler} \tabularnewline
\midrule
\midrule 
Missing & \Checkmark & \Checkmark & \Checkmark & \Checkmark & \Checkmark\tabularnewline
\midrule 
 Hallucination & \Checkmark & \Checkmark & \Checkmark & \Checkmark & \Checkmark\tabularnewline
\midrule 
Size Error & \Checkmark & \Checkmark & - & \Checkmark & \Checkmark\tabularnewline
\midrule 
Split & \Checkmark & - & - & - & -\tabularnewline
\midrule 
Merge & \Checkmark & - & - & \Checkmark & -\tabularnewline
\midrule 
Overlap & \Checkmark & - & - & - & -\tabularnewline
\midrule 
Duplicate & \Checkmark & \Checkmark & - & \Checkmark & \Checkmark\tabularnewline
\midrule 
Misclassification & \Checkmark & \Checkmark & \Checkmark & \Checkmark & \Checkmark\tabularnewline
\bottomrule
\end{tabular}
\end{adjustbox}
\label{tab:Comparison}
\end{table}

\subsection{Comparison with Existing Error Definitions}

The error definitions in this study are designed to be generalizable across document domains and model architectures, providing a unified and extensible foundation for structural error analysis in DLA.
As summarized in Table~\ref{tab:Comparison}, the LED benchmark integrates and standardizes most existing categories while introducing additional document-specific types—such as \textit{Split}, \textit{Merge}, and \textit{Overlap}—that capture hierarchical and compositional structures unique to documents.
These refinements enable LED to diagnose structural inconsistencies overlooked by conventional object-detection analyses, offering a more comprehensive and diagnostic framework for evaluating structural reasoning in document layout understanding.

% \begin{table}[!t]
% \centering
% \caption{Comparison between LED error types and existing research}
% \begin{adjustbox}{max width=\linewidth}
% \begin{tabular}{cccccc}
% \toprule 
% \multirow{2}{*}{Error Type} & \multicolumn{5}{c}{Object Detection Error Type Definition Research}\tabularnewline
% \cmidrule{2-6}
% & LED (Ours) & TIDE & ObjectLab & Loss Inspection~\cite{b7} & DLER~\cite{b9} \tabularnewline
% \midrule
% \midrule 
% Missing & \Checkmark & \Checkmark & \Checkmark & \Checkmark & \Checkmark\tabularnewline
% \midrule 
%  Hallucination & \Checkmark & \Checkmark & \Checkmark & \Checkmark & \Checkmark\tabularnewline
% \midrule 
% Size Error & \Checkmark & \Checkmark & - & \Checkmark & \Checkmark\tabularnewline
% \midrule 
% Split & \Checkmark & - & - & - & -\tabularnewline
% \midrule 
% Metge & \Checkmark & - & - & \Checkmark & -\tabularnewline
% \midrule 
% Overlap & \Checkmark & - & - & - & -\tabularnewline
% \midrule 
% Duplicate & \Checkmark & \Checkmark & - & \Checkmark & \Checkmark\tabularnewline
% \midrule 
% Misclassification & \Checkmark & \Checkmark & \Checkmark & \Checkmark & \Checkmark\tabularnewline
% \bottomrule
% \end{tabular}
% \end{adjustbox}
% \label{tab:Comparison}
% \end{table}

\begin{figure}[!t]
    \centering
    \includegraphics[width=\linewidth]{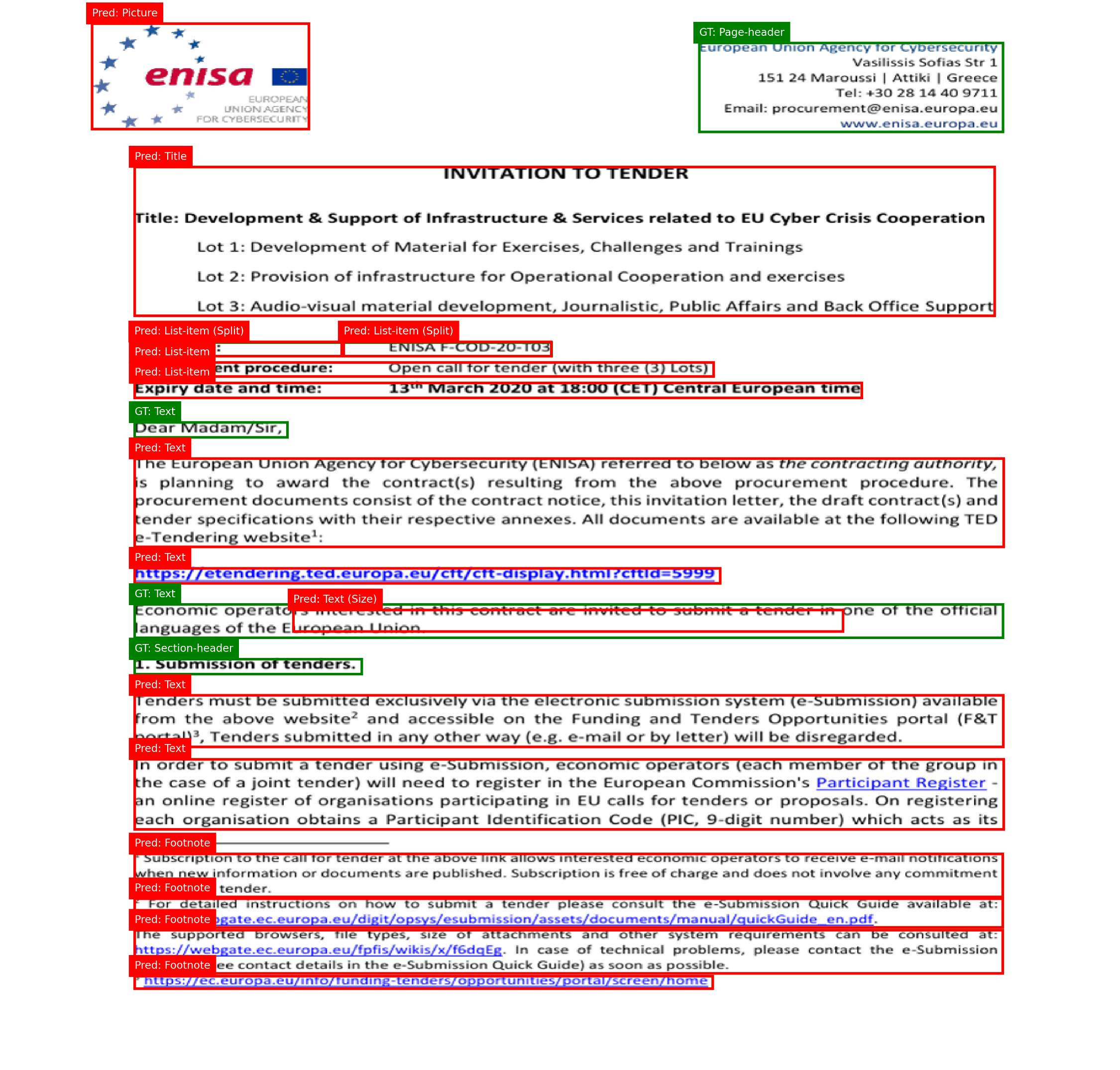}
    \caption{Example from the LED-Dataset showing predicted boxes (red) and ground truth boxes (green). Regions visible only in green indicate missing elements, while red boxes illustrate size discrepancies or split errors within the document layout.}

    \label{fig:sample_image}
\end{figure}

\section{LED-Dataset}

The \textbf{LED-Dataset} is a synthetic benchmark dataset built to enable quantitative diagnosis and comparative evaluation of structural errors in DLA predictions.  
It is constructed upon the test split of DocLayNet, where we apply the proposed error injection algorithm to simulate realistic failure patterns.  
By reproducing the types and frequencies of structural errors commonly observed in existing models, LED-Dataset provides a controlled yet realistic environment for benchmarking the robustness and interpretability of DLA systems.

\subsection{Synthetic Dataset Generation}

We algorithmically inject eight types of structural errors—\textit{Missing, Hallucination, Size Error, Split, Merge, Overlap, Duplicate, and Misclassification}—into document layouts at both the element and document levels. Rather than applying random perturbations, we model the injection probabilities based on error distributions observed from commercial DLA systems, ensuring that the generated samples realistically reproduce real-world error tendencies.

The LED-Dataset includes diverse structural inconsistencies such as Missing, Size Error, and Split.
Fig.~\ref{fig:sample_image}, red boxes represent predicted regions containing structural errors, while green boxes indicate the corresponding ground truth annotations.
Areas visible only in green denote missing elements, whereas discrepancies in red boxes reflect issues such as size errors or split regions.

This example visually demonstrates how different types of structural errors appear within document layouts, providing an intuitive understanding of model behavior in layout prediction.

\begin{figure*}[t]
    \centering
    \includegraphics[width=1.0\textwidth]{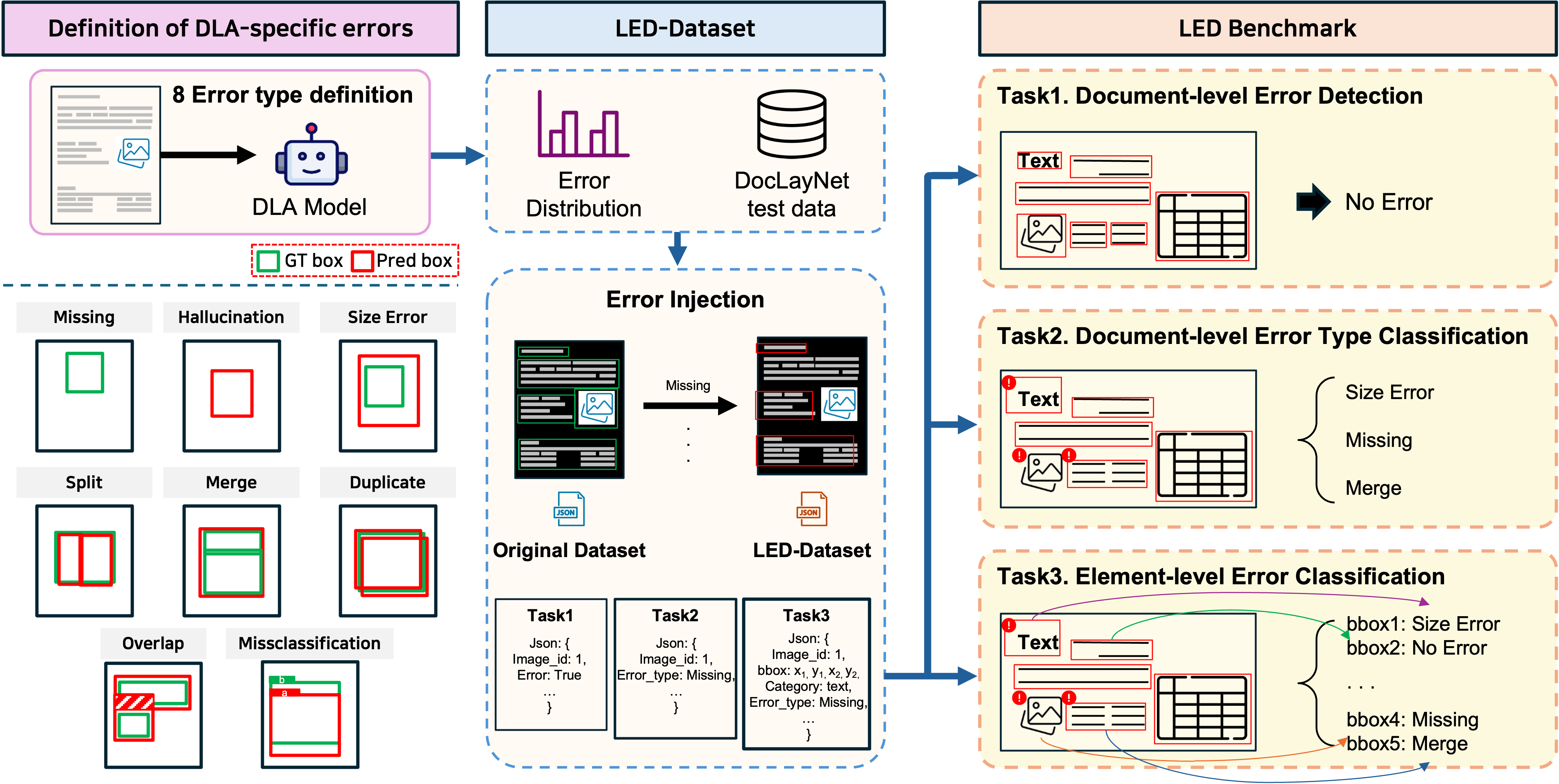}
\caption{
Overview of the proposed \textbf{Layout Error Detection (LED)} framework.
LED defines eight structural error types and builds the \textbf{LED-Dataset} by injecting realistic layout errors into DocLayNet pages.
It evaluates models through three hierarchical tasks—
($T_1$) document-level error detection, 
($T_2$) error type classification, and 
($T_3$) element-level error classification—
enabling fine-grained and explainable assessment of structural robustness in Document Layout Analysis.
}
    \label{fig:overview}
\end{figure*}

\subsection{Raw Data and Annotation Structure}

Each document in the \textit{LED-Dataset} is stored in JSON format and contains the following components:

\begin{itemize}
    \item \textbf{Original Document Image} \\
    The raw page image used for layout analysis.

    \item \textbf{Ground Truth (GT) Annotation} \\
    COCO-style annotation data that includes bounding box coordinates and category IDs for all layout elements.

    \item \textbf{GT Visualization Image} \\
    A rendered image overlaying the ground truth annotations on the original document. 
    It displays the position, ID, and class label of each element for visual inspection.

    \item \textbf{Error Annotation JSON} \\
    A supplementary annotation file indicating the presence of structural errors (binary) 
    and listing the corresponding error types (multi-label) for each document. 
    This file serves as metadata for evaluating error detection and classification tasks.
\end{itemize}

% \begin{figure}[!t]
%     \centering
%     \includegraphics[width=0.9\linewidth]{figure/distribte_error.png}
%     \caption{Distribution of structural error types in LED-Dataset. Missing errors dominate the dataset, followed by Hallucination and Size errors, reflecting real-world error frequency patterns observed in commercial DLA models}
%     \label{fig:distribute}
% \end{figure}

\subsection{Dataset Statistics}

The final \textbf{LED-Dataset} comprises 4{,}996 document images and approximately 70{,}000 layout elements (bounding boxes). Structural errors are injected based on empirical distributions estimated from the predictions of a commercial DLA model, \texttt{maskrcnn\_dit\_base}\footnote{\url{https://huggingface.co/microsoft/dit-base}}. This ensures that the dataset reflects realistic error tendencies frequently observed in deployed systems rather than uniformly sampled noise.

The dataset exhibits a naturally imbalanced error distribution.
Among all error categories, \textit{Missing} occupies the largest proportion (64.89\%), followed by \textit{Hallucination} (14.69\%), \textit{Size Error} (10.97\%), and \textit{Misclassification} (8.77\%).
Less frequent structural inconsistencies such as \textit{Split}, \textit{Merge}, \textit{Overlap}, and \textit{Duplicate} collectively account for less than 2\% of all cases.
This non-uniform distribution reflects the empirical characteristics of real DLA model outputs, offering a realistic basis for evaluating model robustness and structural error handling.

% \begin{figure*}[t]
%     \centering
%     \includegraphics[width=1.0\textwidth]{figure/overview.png}
% \caption{
% Overview of the proposed \textbf{Layout Error Detection (LED)} framework.
% LED defines eight structural error types and builds the \textbf{LED-Dataset} by injecting realistic layout errors into DocLayNet pages.
% It evaluates models through three hierarchical tasks—
% ($T_1$) document-level error detection, 
% ($T_2$) error type classification, and 
% ($T_3$) element-level error classification—
% enabling fine-grained and explainable assessment of structural robustness in Document Layout Analysis.
% }
%     \label{fig:overview}
% \end{figure*}

\section{LED benchmark}

The \textbf{Layout Error Detection (LED)} Benchmark is developed to assess the diagnostic capability of Document Layout Analysis (DLA) models beyond conventional detection accuracy.
It provides a standardized framework to evaluate how effectively models detect, classify, and interpret structural inconsistencies in document layouts.
Fig.~\ref{fig:overview}, LED defines eight DLA-specific structural error types and constructs a synthetic dataset, \textbf{LED-Dataset}, by injecting realistic layout errors into DocLayNet test pages based on empirical error distributions.
This dataset offers a controlled environment for reproducible assessment of model robustness across diverse document structures.
Building on this foundation, LED defines three complementary tasks—$T_1$, $T_2$, and $T_3$—to evaluate document-level error detection, document-level error type classification, and element-level error classification.

\subsection{Task Definition}

The \textbf{LED benchmark} defines three hierarchical tasks—$T_1$, $T_2$, and $T_3$—that evaluate a model’s ability to \textit{detect} and \textit{classify} structural layout errors in document images.  
These tasks go beyond conventional accuracy-based metrics by assessing how well a Document Layout Analysis (DLA) model understands the logical and spatial consistency of document structures.  
Importantly, LED is designed to support \textbf{post-correction of DLA outputs} and \textbf{model weakness diagnosis}, providing practical insights for improving structural robustness in document understanding systems.

\subsubsection*{\textbf{$T_1$: Document-level Error Detection}}
This task evaluates whether a document’s predicted layout contains any structural errors.
Formulated as a binary classification problem, it assesses a model’s structural awareness—its ability to judge if a document remains logically coherent beyond geometric alignment (e.g., IoU).
Each input includes a document image and its predicted layout (JSON or overlay), and the output is a binary label indicating the presence of structural inconsistencies.
Performance is measured using \textit{Accuracy}, providing a fast and practical indicator for large-scale quality control and identifying cases requiring post-correction.

\subsubsection*{\textbf{$T_2$: Document-level Error Type Classification}}
This task identifies which structural error types exist within a document.
Defined as a multi-label classification problem, each document is represented by an eight-dimensional binary vector corresponding to predefined error types (e.g., Missing, Merge, Hallucination).
Using the same input as $T_1$, the model must reason across layout components to infer dominant error categories.
Evaluation uses \textit{F1-score} (micro and macro averages) to measure how effectively models distinguish among error patterns.
$T_2$ provides mid-level interpretability, helping reveal systematic weaknesses and informing post-correction or model refinement.

\subsubsection*{\textbf{$T_3$: Element-level Error Type Classification}}
This task provides fine-grained analysis by classifying structural errors for each predicted bounding box.
Each element is labeled with one of eight error types or “None” for correct predictions, testing the model’s ability to recognize local structural inconsistencies.
Inputs consist of a document image with predicted and ground-truth boxes, and performance is evaluated using \textit{macro-averaged F1-score}.
$T_3$ enables detailed diagnosis of model behavior—identifying which layout elements (e.g., text, tables, figures) are most error-prone and offering insights for targeted correction strategies.

These tasks establish a hierarchical framework for diagnosing structural reasoning in DLA models.

\subsection{Prompting Configuration}

To investigate how contextual information influences a model’s ability to detect and classify structural layout errors,  
we introduce three prompting configurations ($P_1$–$P_3$) that vary in the type and richness of the input provided to the model.  
These configurations are applied consistently across all tasks ($T_1$–$T_3$) to analyze the relationship between input context and structural reasoning performance.  
Such comparative settings also serve as diagnostic probes for understanding how multimodal models leverage visual and structural cues—providing practical insights for DLA post-correction and model refinement.

\begin{itemize}
    \item \textbf{$P_1$: Page Image \& Prediction JSON}  
    The model receives both the document image and the predicted layout in JSON format.  
    This configuration allows reasoning over visual appearance and structured spatial data simultaneously, simulating scenarios where both modalities are available for post-correction.

    \item \textbf{$P_2$: Page Image with Visualized Bounding Boxes}  
    Only the document image with predicted bounding boxes overlaid is provided.  
    The model must rely solely on visual cues to infer structural irregularities, serving as a baseline to measure purely visual reasoning capability.

    \item \textbf{$P_3$: Page Image \& Visualized Bounding Boxes \& JSON}  
    Both the visualized layout and the structured JSON are jointly provided, offering the richest contextual information.  
    This configuration evaluates the model’s multimodal reasoning ability and its effectiveness in integrating visual and structured representations for structural error diagnosis.
\end{itemize}

Overall, these prompting configurations enable systematic analysis of how different input modalities contribute to structural understanding,  
revealing which information sources most effectively support accurate detection, classification, and post-correction of layout errors.

\section{Experimental Setup}

We evaluated the \textbf{LED benchmark} across three diagnostic tasks ($T_1$–$T_3$) using 4{,}996 samples from the \textbf{LED-Dataset}.
A diverse set of multimodal models of varying architectures and scales were tested under identical zero-shot conditions to assess inherent structural reasoning capability.
All inputs followed three prompting configurations ($P_1$–$P_3$), and model outputs were evaluated using standardized LED scripts with \textit{Accuracy} for $T_1$ and \textit{F1-score} for $T_2$–$T_3$.
Results were analyzed to measure each model’s robustness and capability to interpret structural inconsistencies in document layouts.

\begin{table*}[!t]
    \caption{LED benchmark performance by model across tasks and prompting. \textbf{Bold} indicates the best-performing method per task; \underline{\textbf{bold\&underline}} highlights the overall best model–prompt pair.
}
    \centering
    \begin{tabular}{cr@{\extracolsep{0pt}.}lr@{\extracolsep{0pt}.}lr@{\extracolsep{0pt}.}l|ccc|ccc}
    \hline 
    \noalign{\vskip2dd}
    \multicolumn{1}{c}{} & \multicolumn{6}{c|}{Task1} & \multicolumn{3}{c|}{Task2} & \multicolumn{3}{c}{Task3}\tabularnewline
    \cline{2-13}
    \noalign{\vskip2dd}
    Model & \multicolumn{2}{c}{Prompt 1} & \multicolumn{2}{c}{Prompt 2} & \multicolumn{2}{c|}{Prompt 3} & Prompt 1 & Prompt 2 & Prompt 3 & Prompt 1 & Prompt 2 & Prompt 3\tabularnewline
    \cline{2-13}
    \noalign{\vskip2dd}
     & \multicolumn{6}{c|}{Accuracy} & \multicolumn{3}{c|}{F1-Score} & \multicolumn{3}{c}{F1-Score}\tabularnewline
    \hline 
    \noalign{\vskip2dd}
    \multirow{1}{*}{GPT-4o} & \textbf{0}&\textbf{597} & 0&567 & 0&591 & \textbf{0.287} & 0.085 & 0.235 & \textbf{0.066} & 0.012 & 0.044\tabularnewline
    \noalign{\vskip2dd}
    \multirow{1}{*}{GPT-4o-mini} & 0&538 & 0&460 & \textbf{0}&\textbf{560} & \textbf{0.323} & 0.009 & 0.156 & \textbf{0.159} & 0.034 & 0.104\tabularnewline
    \noalign{\vskip2dd}
    \multirow{1}{*}{Gemini 2.5 Pro} & \textbf{\underline{0}}&\textbf{\underline{636}} & 0&626 & 0&603 & \textbf{\underline{0.598}} & 0.490 & 0.580 & \textbf{\underline{0.443}} & 0.369 & 0.407\tabularnewline
    \noalign{\vskip2dd}
    \multirow{1}{*}{Gemini 2.5 Flash} & 0&610 & 0&586 & \textbf{0}&\textbf{614} & \textbf{0.432} & 0.372 & 0.414 & \textbf{0.333} & 0.266 & 0.284\tabularnewline
    \noalign{\vskip2dd}
    \multirow{1}{*}{Gemini 2.5 Flash Lite} & 0&421 & \textbf{0}&\textbf{435} & 0&432 & \textbf{0.334} & 0.229 & 0.305 & \textbf{0.127} & 0.056 & 0.117\tabularnewline
    \hline 
    \noalign{\vskip2dd}
    \multirow{1}{*}{DeepSeek V3} & \textbf{0}&\textbf{458} & 0&406 & 0&456 & \textbf{0.127} & 0.011 & 0.095 & 0.133 & 0.114 & \textbf{0.147}\tabularnewline
    \noalign{\vskip2dd}
    \multirow{1}{*}{Llama 4 Maverick} & \textbf{0}&\textbf{476} & 0&435 & 0&468 & 0.124 & 0.040 & 0.101 & \textbf{0.075} & 0.005 & 0.064\tabularnewline
    \noalign{\vskip2dd}
    \multirow{1}{*}{Llama 4 Scout} & 0&461 & 0&468 & \textbf{0}&\textbf{515} & \textbf{0.099} & 0.013 & 0.071 & \textbf{0.002} & 0.001 & 0.001\tabularnewline
    \hline 
    \end{tabular}
    \label{tab:overall_perf}
\end{table*}

\subsection{Model Pool \& Size}

Our model pool includes both closed-source commercial APIs and open-weight models. In total, we evaluate eight models:

\begin{itemize}
    \item \textbf{GPT}: \textit{GPT-4o}, \textit{GPT-4o-mini}~\cite{openai2024gpt4o}
    \item \textbf{Gemini}: \textit{Gemini 2.5 Pro}, \textit{Gemini 2.5 Flash}, \textit{Gemini 2.5 Flash Lite}~\cite{google2024gemini}
    \item \textbf{DeepSeek}: \textit{DeepSeek V3}~\cite{deepseek2024v3}
    \item \textbf{LLaMA}: \textit{LLaMA 4 Maverick}, \textit{LLaMA 4 Scout}~\cite{meta2024llama4}
\end{itemize}

These models span a variety of architectural families, training paradigms, and parameter scales, enabling a broad comparison across the structural error detection spectrum.
This diverse selection is intended to systematically compare how model family and scale impact performance on LED tasks.

\subsection{Implementation \& API Setting}

All models were accessed through a unified API interface via the OpenRouter platform\footnote{\url{https://openrouter.ai/}}. To ensure reproducibility and fairness, decoding parameters were fixed across all runs: temperature = 1.0, top-p = 1.0, and repetition penalty = 1.0. Prompt formats were kept consistent across all models. Differences in maximum input length, response latency, and tokenizer behavior were noted and considered as auxiliary factors during result interpretation.

\section{Results and Analysis}
We evaluate the proposed LED benchmark to examine how effectively current multimodal models can detect and interpret structural layout errors in documents. 
The analysis covers overall model accuracy, scalability by size and family, task-wise behavior across $T_1$–$T_3$, 
and the influence of input composition ($P_1$–$P_3$). 
Through these experiments, LED reveals clear performance gaps between general layout recognition and fine-grained structural understanding.

\subsection{Overall Performance}
Among all evaluated models, \textit{Gemini 2.5 Pro} and \textit{Gemini 2.5 Flash} consistently achieved the highest and most stable performance across all LED tasks.  
Their strong F1-scores on $T_2$ and $T_3$ (0.33–0.60 overall, up to 0.60 on $T_2$ and 0.44 on $T_3$) indicate superior ability not only to detect but also to interpret fine-grained structural errors in document layouts.  
In contrast, the \textit{GPT-4o} family exhibited strong performance on $T_1$ (binary error detection) but showed a clear drop in accuracy and F1-scores on $T_2$ and $T_3$, suggesting limited capacity for fine-grained structural reasoning.  

A full summary of the quantitative results for all models, tasks, and input configurations ($P_1$–$P_3$) is provided in Table~\ref{tab:overall_perf}.

\subsection{Model-size Trends}
Model scale influenced performance, but the effect varied by architecture.  
In the \textit{GPT} family, the smaller \textit{GPT-4o-mini} even outperformed the larger variant in $T_3$ (F1: 0.159 vs.\ 0.066), implying that smaller models may retain stronger specialization in specific error patterns.  
Conversely, the \textit{Gemini} models showed a clear positive correlation between size and performance: F1-scores for $T_2$ and $T_3$ steadily increased from Flash Lite $\rightarrow$ Flash $\rightarrow$ Pro (e.g., 0.229 $<$ 0.372 $<$ 0.490), suggesting that larger Gemini variants leverage their enhanced multimodal fusion capabilities for improved structural reasoning.  
Meanwhile, \textit{DeepSeek} and \textit{LLaMA 4} consistently underperformed regardless of scale, indicating that architectural limitations—rather than parameter count—may be the dominant factor affecting their LED performance.

\subsection{Model-family Trends}
Performance patterns varied significantly across model families.  
The \textit{Gemini} series demonstrated the most stable and high-performing results across all three tasks, effectively integrating visual and semantic cues for structural reasoning.  
In contrast, the \textit{GPT} family showed large variance depending on the input setting: while strong in $T_1$, their performance degraded sharply in $T_2$ and $T_3$.  
The \textit{DeepSeek} and \textit{LLaMA 4} families showed the weakest results overall, with low accuracy and near-zero F1-scores in element-level classification ($T_3$), suggesting insufficient alignment with document-structural objectives.  
Overall, Gemini models appear best suited for multimodal document reasoning, while GPT models exhibit partial interpretability, and DeepSeek/LLaMA remain poorly aligned with LED-style evaluation.

\subsection{Task-wise Trends ($T_1$ vs. $T_2$ vs. $T_3$)}
The LED benchmark’s three hierarchical tasks—$T_1$ (error existence), $T_2$ (error type classification), and $T_3$ (element-level classification)—evaluate progressively deeper structural understanding.  
Most models achieved high accuracy on $T_1$, confirming that detecting whether a document contains any structural error is relatively easy.  
However, performance dropped sharply on $T_2$ and even further on $T_3$, where distinguishing and localizing fine-grained error types is required.  
This trend highlights that while most models can sense the presence of anomalies, few can accurately diagnose the type or location of structural inconsistencies.  
The results demonstrate that LED effectively differentiates between general detection ability and deeper interpretive reasoning about document structure.

\begin{figure}[!t]
    \centering
    \includegraphics[width=1.0\linewidth]{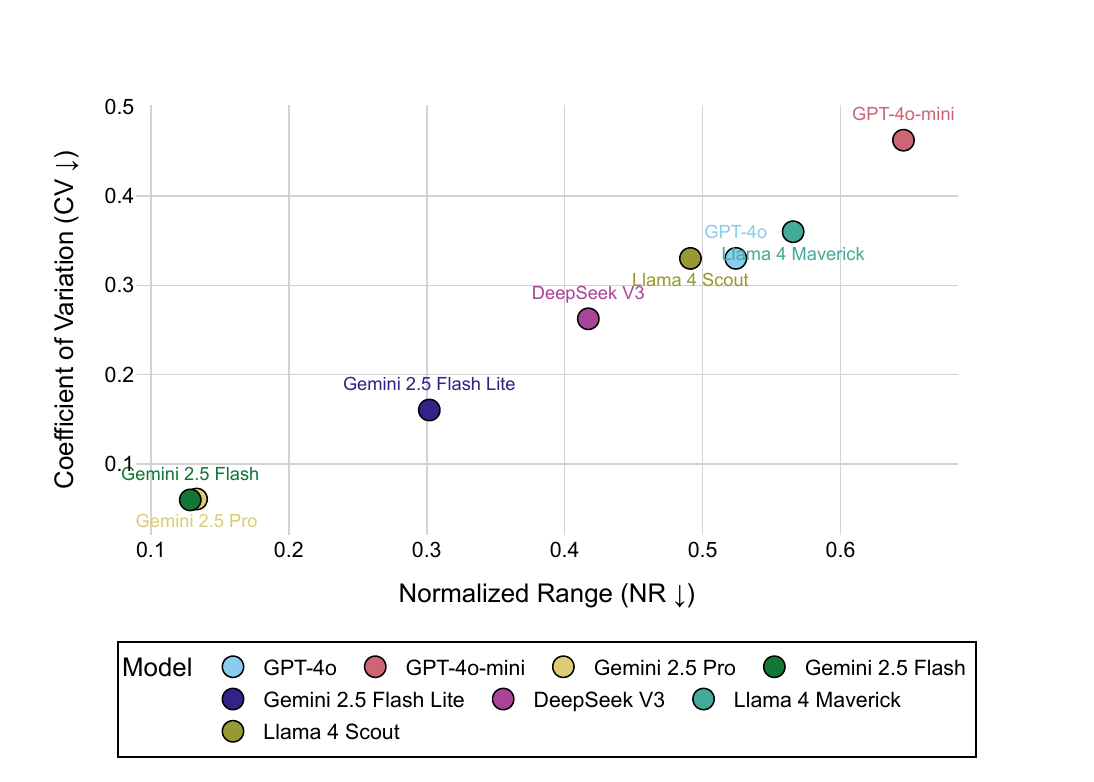}
    \caption{Prompting-wise robustness across models (lower CV/NR = higher stability)}
    \label{fig:prompt}
\end{figure}

\section{Prompting-wise Analysis}

Prompt configuration strongly influenced model performance and stability across all LED tasks.  
Among the three settings, \textbf{$P_1$} (page image with prediction JSON) consistently yielded the best accuracy, while \textbf{$P_2$} (page image with overlaid boxes) showed the weakest results.  
\textbf{$P_3$} (image \& boxes \& JSON) generally ranked between the two, indicating that structured textual–spatial representations contribute more to reasoning than purely visual cues.  
These trends were especially evident within the \textit{Gemini} family, where richer multimodal inputs improved performance, whereas \textit{GPT} models sometimes degraded when additional modalities were introduced, suggesting architectural differences in multimodal signal fusion.

To further quantify robustness to prompt variation, we measured the \textit{Coefficient of Variation (CV)} and \textit{Normalized Range (NR)} across prompts for each model and task.  
Lower CV and NR indicate greater stability and prompt invariance.  
Fig.~\ref{fig:prompt}, \textit{Gemini} models achieved the lowest CV/NR values, confirming strong multimodal alignment and consistent reasoning.  
In contrast, \textit{GPT} models exhibited higher variance—particularly under $P_2$—while open-weight models (\textit{LLaMA 4}, \textit{DeepSeek V3}) showed the largest fluctuations, implying weaker layout-structural robustness.  
Overall, these results demonstrate that prompt design significantly impacts multimodal document understanding and that model architectures differ in their ability to integrate heterogeneous input modalities.

\begin{figure}[!t]
    \centering
    \includegraphics[width=0.9\linewidth]{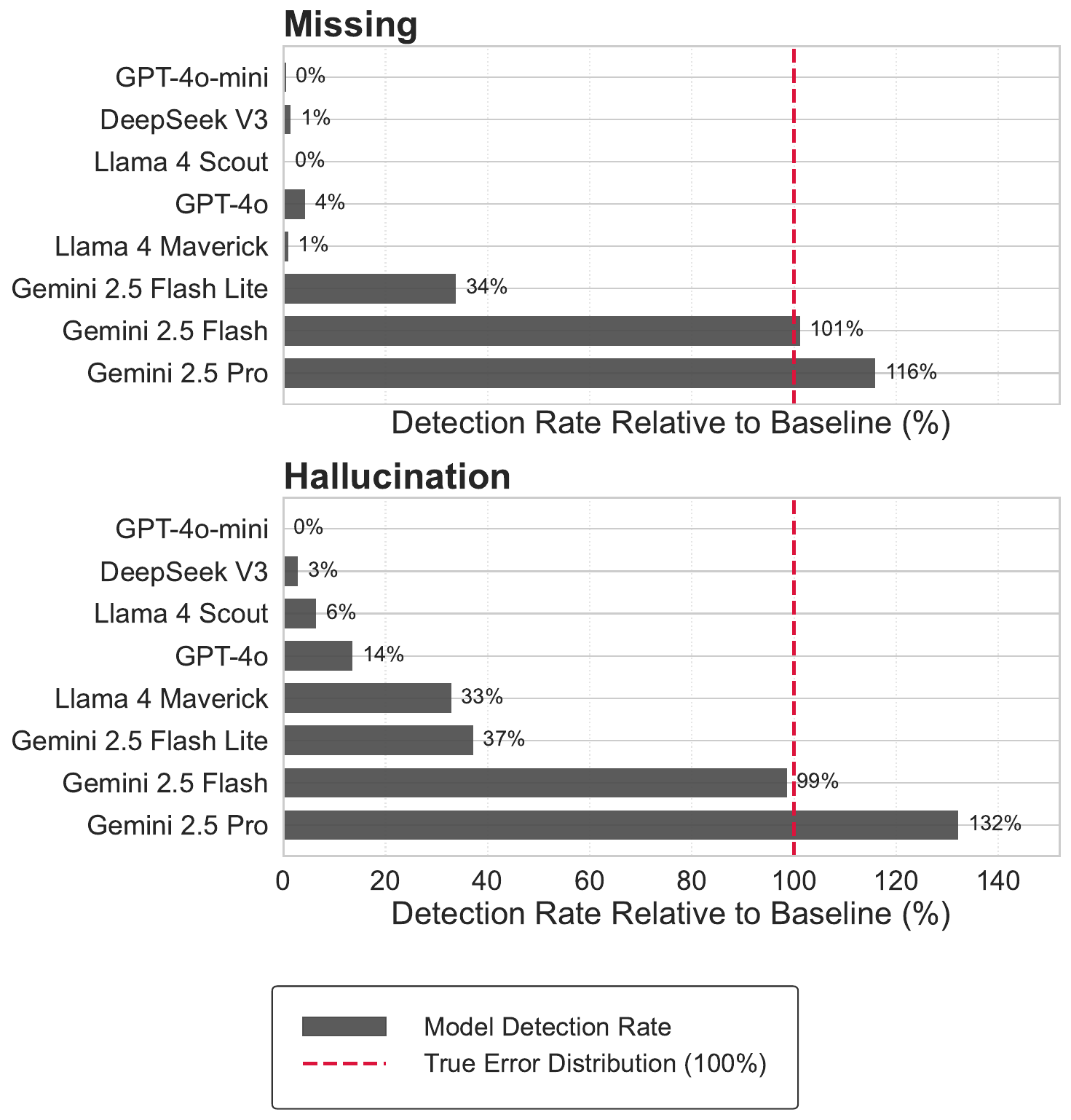}
    \caption{Detection rates for the two most frequent error types in LED ($T_2$ task). Red dashed lines indicate true error distributions.}
    \label{fig:error}
\end{figure}

\subsection{Error-type Trends}
To further analyze model behavior, we examined performance on the two most frequent error categories in the LED dataset: \textit{Missing} and \textit{Hallucination}.  
Fig.~\ref{fig:error}, presents the detection rates for these error types under $T_2$, where the red dashed line represents the true distribution of each error.  
Models whose detection rates align closely with these baselines demonstrate higher reliability.  
The \textit{Gemini} family models showed balanced proficiency across both error types, while most others—particularly smaller GPT and DeepSeek variants—struggled to capture Missing errors.  
Interestingly, in Hallucination detection, \textit{LLaMA 4 Maverick} performed comparably to \textit{Gemini 2.5 Flash}, indicating that some models may partially compensate for weak overall accuracy with specialized strength in certain error categories.

\section{Discussion}

% This study presents \textbf{Layout Error Detection (LED)}, a diagnostic benchmark for evaluating structural consistency in Document Layout Analysis (DLA) models.  
% Unlike traditional metrics (e.g., IoU, mAP) focused on geometric overlap, LED identifies semantically interpretable error types—\textit{Missing}, \textit{Merge}, and \textit{Hallucination}—for explainable assessment of model robustness.  
% By defining eight error types and building a realistic synthetic dataset, LED bridges quantitative accuracy with qualitative structural understanding.

\textbf{LED} is a benchmark for evaluating structural consistency in Document Layout Analysis (DLA) models.  
Unlike overlap-based metrics (e.g., IoU, mAP), LED defines semantically interpretable error types and a realistic synthetic dataset for explainable assessment of model robustness.

Experiments across eight multimodal models reveal three key findings.  
\textbf{(1) Model Family:} \textit{Gemini} models achieved the best overall results, while \textit{GPT} models excelled in coarse detection but struggled with fine-grained structural reasoning.  
Open-weight models (\textit{LLaMA 4}, \textit{DeepSeek}) showed limited performance, indicating current architectures are not optimized for layout reasoning.  
\textbf{(2) Input Sensitivity:} Model robustness varied with prompt complexity—multimodal inputs improved performance, but overly complex prompts degraded reasoning, emphasizing the need for prompt-adaptive fusion.  
\textbf{(3) Error-type Asymmetry:} Models handled hallucination errors well but failed on \textit{Missing} and \textit{Size} types, exposing weak spatial and contextual reasoning.

% LED’s limitations include its independence from baseline detection quality, its focus on core structural errors without higher-order semantics, and its domain-limited dataset.  
% Future work will expand LED to multi-domain and multilingual settings, integrating it with DLA correction pipelines for deeper insights into document-structural reasoning.

LED’s limitations include the exclusion of baseline detection quality as well as the use of a domain-limited dataset.

\section{Conclusion}

This study presents \textbf{Layout Error Detection (LED)}, a diagnostic benchmark for evaluating structural reasoning in Document Layout Analysis (DLA) models.  
LED defines eight error types, employs a rule-based error injection framework, and builds the synthetic \textbf{LED-Dataset} reflecting realistic structural failures.  
Through three hierarchical tasks—document-level detection, error-type classification, and element-level diagnosis—LED enables explainable evaluation beyond geometric accuracy.

Experimental results reveal three insights:
(1) \textbf{Model-family differences:} \textit{Gemini} shows balanced multimodal reasoning, \textit{GPT} performs well in detection but struggles with structure, and \textit{LLaMA}/\textit{DeepSeek} remain limited.  
(2) \textbf{Prompt sensitivity:} Multimodal inputs enhance some models but destabilize language-centric ones.  
(3) \textbf{Error-type asymmetry:} Models detect hallucinations well but miss spatial or missing errors.

\textbf{LED} provides a means to verify whether models go beyond detection to genuinely understand and reason over document structures.

% \section*{Acknowledgment}

% The preferred spelling of the word ``acknowledgment'' in America is without 
% an ``e'' after the ``g''. Avoid the stilted expression ``one of us (R. B. 
% G.) thanks $\ldots$''. Instead, try ``R. B. G. thanks$\ldots$''. Put sponsor 
% acknowledgments in the unnumbered footnote on the first page.

\end{document}